# OMGs: A multi-agent system supporting MDT decision-making across the ovarian tumour care continuum

Yangyang Zhang[1,2,†], Zilong Wang[3,†], Jianbo Xu[4,†], Yongqi Chen[1,2,†], Chu Han[5,6], Zhihao Zhang[2,7], Shuai Liu[2,8], Hui Li[2,9], Huiping Zhang[10], Ziqi Liu[1,2], Jiaxin Chen[1,2], Jun Zhu[1,2], Zheng Feng[1,2], Hao Wen[1,2], Xingzhu Ju[1,2], Yanping Zhong[1,2], Yunqiu Zhang[11], Jie Duan[12], Jun Li[12], Dongsheng Li[3], Weijie Wang[4], Haiyan Zhu[11], Wei Jiang[12], Xiaohua Wu[1,2,*], Shuo Wang[13,*], Haiming Li[2,7,*], Qinhao Guo[1,2,*]

[1]Department of Gynecologic Oncology, Fudan University Shanghai Cancer Center, Fudan University, Shanghai 200032, China.

[2]Department of Oncology, Shanghai Medical College, Fudan University, Shanghai 200032, China.

[3]Microsoft Research Asia, Shanghai 200232, China.

[4]Northern Jiangsu People's Hospital, Yangzhou 225001, China.

[5]Medical Research Institute, Guangdong Provincial People's Hospital (Guangdong Academy of Medical Sciences), Southern Medical University, Guangzhou 510080, China.

[6]Department of Radiology, Guangdong Provincial People's Hospital (Guangdong Academy of Medical Sciences), Southern Medical University, Guangzhou 510080, China

[7]Department of Radiology, Fudan University Shanghai Cancer Center, Fudan University, Shanghai 200032, China.

[8]Department of Nuclear Medicine, Fudan University Shanghai Cancer Center, Fudan University, Shanghai 200032, China.

[9]Department of Pathology, Fudan University Shanghai Cancer Center, Fudan University, Shanghai 200032, China.

[10]Department of Obstetrics and Gynecology, Taizhou First People's Hospital, Taizhou 318020, China

[11]Department of Gynecology, Shanghai Key Laboratory of Maternal Fetal Medicine, Shanghai Institute of Maternal-Fetal Medicine and Gynecologic Oncology, Shanghai First Maternity and Infant Hospital, School of Medicine, Tongji University, Shanghai 200092, China.

[12]Institute of Obstetrics and Gynecology, Hospital of Obstetrics and Gynecology, Fudan University, Shanghai 200080, China.

[13]Digital Medical Research Centre, School of Basic Medical Sciences, Fudan University, Shanghai 200032, China. Shanghai Key Laboratory of MICCAI, Shanghai 200032, China.

[†]These authors contributed equally.

*Corresponding author.

E-mail addresses: wu.xh@fudan.edu.cn (Xiaohua Wu), shuowang@fudan.edu.cn (Shuo Wang), lihaiming0109@163.com (Haiming Li), guoqinhao911@163.com (Qinhao Guo).

**Abstract**

Ovarian tumour management has increasingly relied on multidisciplinary tumour board (MDT) deliberation to address treatment complexity and disease heterogeneity. However, most patients worldwide lack access to timely expert consensus, particularly in resource-constrained centres where MDT resources are scarce or unavailable. Here we present OMGs (Ovarian tumour Multidisciplinary intelligent aGent System), a multi-agent AI framework where domain-specific agents deliberate collaboratively to integrate multidisciplinary evidence and generate MDT-style recommendations with transparent rationales. To systematically evaluate MDT recommendation quality, we developed SPEAR (Safety, Personalization, Evidence, Actionability, Robustness) and validated OMGs across diverse clinical scenarios spanning the care continuum. In multicentre re-evaluation, OMGs achieved performance comparable to expert MDT consensus ($4.45 \pm 0.30$ versus $4.53 \pm 0.23$), with higher Evidence scores (4.57 versus 3.92). In prospective multicentre evaluation (59 patients), OMGs demonstrated high concordance with routine MDT decisions. Critically, in paired human–AI studies, OMGs most substantially enhanced clinicians' recommendations in Evidence and Robustness, the dimensions most compromised when multidisciplinary expertise is unavailable. These findings suggest that multi-agent deliberative systems can achieve performance comparable to expert MDT consensus, with potential to expand access to specialized oncology expertise in resource-limited settings.

**Introduction**

Ovarian malignancies, particularly epithelial ovarian cancer, remain a leading cause of mortality among gynaecologic malignancies.[1] Late-stage diagnosis, substantial inter- and intratumoural heterogeneity, and frequent relapse drive repeated high-stakes decisions across the disease course.[2] Optimized ovarian tumour management requires coordinated expertise from surgery, medical oncology, radiology, pathology, and molecular testing, and is shaped by treatment line, platinum sensitivity, evolving biomarkers (e.g., BRCA, HRD), and serial reassessments over time.[3-6] Multidisciplinary tumour boards (MDTs) are widely used to augment decision-making, but routine case review is challenging to sustain in high-volume settings with fragmented longitudinal records and limited transparency of decision rationale.[7-9] These challenges highlight the urgent need for scalable, accurate, and interpretable decision-support tools for ovarian tumours, an area where recent advances in multi-agent systems for healthcare (MASH) may be well suited to these demands.[10]

Developing artificial intelligence systems for ovarian tumour decision support presents several inherent challenges.[11] First, the task is multidisciplinary, requiring role-specific reasoning across specialties and reconciliation when interpretations diverge.[7,12] Second, the decision space shifts across clinical scenes over the care continuum, spanning primary management, histology-driven pathways, platinum-resistant relapse, platinum-sensitive relapse, and event-driven reassessment.[13] Third, inputs are longitudinal and possibly incomplete, distributed across time-ordered reports with missing data, uncertainty, and discordant assessments.[14] Fourth, clinical deployment requires transparency and traceability, linking recommendations to verifiable patient-specific evidence and making uncertainty and reassessment explicit to support accountability and audit.[15] In this context, single-agent large language model (LLM) assistants may be insufficient to ensure that recommendations remain consistently verifiable and traceable across incomplete and evolving longitudinal records.[16,17]

Recent advances in agentic large language model (LLM) systems have opened new avenues for complex oncology decision support by coordinating specialized sub-agents with explicit task boundaries.[18-21] Such systems naturally fit the MDTs and enable seamless integration of multimodal clinical data, external knowledge bases, and case repositories through coordinated tool use, while reconciling conflicting perspectives via structured reasoning and evidence attribution.[22-24] Compared with single-agent approaches, a multi-agent design may better structure role-specific reasoning and facilitate evidence-grounded reporting, which is hypothesized to improve auditability under fragmented records.[12,25,26] These properties match the operational requirements of tumour boards, where accountability and reassessment are integral to safe decision-making.

Here, we developed OMGs (Ovarian tumour Multidisciplinary intelligent aGent System), an LLM-powered multi-agent MDT decision-support framework tailored to ovarian tumours. OMGs resembles key MDT specialties (MDT Chair, medical oncology, radiology, pathology, and nuclear medicine) under role-specific constraints. Structured patient information with time-ordered reports is presented to OMGs for generating standardized MDT-style recommendations with traceable evidence support and explicit reassessment triggers. We evaluated OMGs in retrospective and prospective multicentre settings using a prespecified Likert scale covering Safety, Personalization, Evidence strength and traceability, Actionability, and Robustness (SPEAR). We compared OMGs against clinician baselines, state-of-the-art AI models, and routine MDT conclusions. We also conducted paired human–AI studies to quantify incremental gains when clinicians were supported by OMGs, including resident physicians at tertiary cancer centres and senior physicians from non-tertiary hospitals.

## Results

### System overview

OMGs is an LLM-powered multi-agent system for ovarian tumour MDT decision support (**Fig. 1a**). It implements a two-tier architecture comprising (i) an agent orchestrator that coordinates five specialty-specific LLM agents through structured MDT deliberation; and (ii) agent servers that support clinical data extraction, context assembly, evidence retrieval, report selection, and provenance tracking.

Upon receiving a patient case with longitudinal clinical records, the agent orchestrator coordinates five specialty agents (chair, oncologist, radiologist, pathologist, and nuclear medicine physician), each equipped with domain-specific expertise and curated clinical data, to analyse the case from their respective domains. Each agent retrieves relevant evidence from clinical practice guidelines, biomedical literature, and trial registries tailored to their specialty. The agents then engage in multi-round, evidence-backed deliberation, where each specialist can challenge, clarify, and refine the interpretations of others to reach consensus, mirroring the dynamics of a real-world MDT meeting. The system outputs a structured decision summary with Final Assessment, Treatment Strategy, and Change Triggers, all linked to traceable evidence sources (**Extended Data Fig. 1**). Additional implementation details are provided in the Methods.

### Study design, clinical scenes and cohorts

The system was evaluated in retrospective and prospective multicentre cohorts spanning five prespecified MDT clinical scenes along the ovarian tumour care continuum: primary management, histology-driven pathways, platinum-resistant relapse, platinum-sensitive relapse, and event-driven reassessment (**Fig. 1b and Supplementary Table S1**). Evaluation comprised four sequential phases (**Fig. 2a and Extended Data Fig. 3**): single-centre retrospective assessment (Phase I), multicentre retrospective re-MDT evaluation (Phase II; independent MDT re-review blinded to original decisions), prospective comparison with real-world MDT conclusions (Phase III), and assessment of human–AI collaboration in routine workflows (Phase IV). System outputs were evaluated using a prespecified five-dimension SPEAR rubric, with each dimension rated on a 1–5 Likert scale, assessing Safety, Personalization, Evidence strength and traceability, Actionability, and Robustness (**Supplementary Table S2**).

Across participating centres, de-identified real-world clinical records of ovarian tumours were screened for eligibility (**Extended Data Fig. 2**). After exclusion of patients with incomplete clinical information or incomplete MDT documentation, a total of 304 retrospective patients were included from three centres: Fudan University Shanghai Cancer Center (FUSCC; $n = 253$), Northern Jiangsu People's Hospital (NJPH; $n = 30$), and Taizhou First People's Hospital (TFPH; $n = 21$) (**Table 1**). The median age was 55.0 years (IQR, 47.0–62.0). Epithelial ovarian carcinoma accounted for the majority of cases (82.2%), with additional representation from borderline epithelial tumours, germ cell tumours, sex cord–stromal tumours, sarcomas, neuroendocrine neoplasms, and other rare aggressive entities. Baseline characteristics were broadly comparable across centres, whereas FIGO stage and case composition varied, reflecting real-world referral heterogeneity (**Table 1 and Supplementary Table S3**).

Prospective evaluation included 59 consecutive MDT patients from three centres: a prospective cohort at Fudan University Shanghai Cancer Center (FUSCC; $n = 39$), Fudan University Obstetrics and Gynecology Hospital (FOGH; $n = 10$), and Shanghai First Maternity and Infant Hospital (SFMIH; $n = 10$) (**Extended Data Table 1**). The median age was 56.0 years (IQR, 50.0–65.0), and the distribution of histologic families was similar to that observed in the retrospective cohort. Differences in FIGO stage and clinical scene distribution across centres were again observed, consistent with routine clinical practice.

**Phase I: Single-centre retrospective benchmarking**

We first conducted a blinded single-centre retrospective benchmarking study using 253 consecutive ovarian tumour cases from FUSCC to quantify the contributions of multi-agent deliberation and progressively enriched information contexts (**Fig. 2**). Using the same final chair summarization module and the same schema-constrained MDT output template, we compared OMGs against three single-agent chair baselines that differed only in information input. CHAIR-R used the schema-normalized case representation only. CHAIR-E added external evidence retrieval while keeping the same structured case inputs. CHAIR-D further added a case-specific evidence dossier comprising the full underlying clinical reports, including pathology, imaging, laboratory and genomic reports, together with retrieved evidence and an optional list of candidate clinical trials. This design enables attribution of gains from evidence augmentation (CHAIR-E versus CHAIR-R), dossier-based enrichment of clinical context (CHAIR-D versus CHAIR-E), and multi-specialty deliberation beyond the most informed single-agent baseline (OMGs versus CHAIR-D).

Across all patient cases, OMGs achieved the highest mean scores across all five SPEAR dimensions (**Fig. 2b**). Ratings were consistent across the three senior gynaecologic oncology experts (**Extended Data Fig. 4 and Supplementary Table S4**). OMGs achieved mean scores of $4.36 \pm 0.62$ for Safety, $4.26 \pm 0.64$ for Personalization, $4.19 \pm 0.64$ for Evidence, $4.18 \pm 0.77$ for Actionability, and $4.37 \pm 0.55$ for Robustness, exceeding the corresponding scores of CHAIR-D ($4.12 \pm 0.60$, $4.11 \pm 0.62$, $4.04 \pm 0.60$, $3.92 \pm 0.76$, $3.72 \pm 0.56$, respectively) as well as those of CHAIR-E and CHAIR-R (**Supplementary Table S5**). This pattern was consistent across all dimensions, with the largest absolute gaps in Robustness and Evidence.

The distribution of high-scoring recommendations further distinguished OMGs from single-agent baselines (**Fig. 2c**). The proportion of OMGs outputs achieving scores ≥ 4 reached 92% for Safety, 89% for Personalization, 87% for Evidence, 86% for Actionability, and 97% for Robustness. In contrast, CHAIR-R achieved high-score proportions of 30%, 42%, 23%, 32%, and 20%, respectively, indicating substantial differences not only in mean performance but also in the proportion of cases achieving high scores. When overall decision quality was evaluated using a safety-gated aggregation, where scores across dimensions were capped by the Safety score when it fell below a prespecified threshold (see **Methods**), OMGs again demonstrated the highest performance across all cases. The mean overall SPEAR score of OMGs was 4.27 ± 0.31, compared with 3.98 ± 0.29 for CHAIR-D, 3.31 ± 0.46 for CHAIR-E, and 2.84 ± 0.58 for CHAIR-R (**Supplementary Table S5**).

Scene-stratified analyses showed that the relative performance of OMGs was consistent across all five clinical scenes (**Fig. 2d**). OMGs maintained mean overall SPEAR scores above 4.24 in each scene, whereas CHAIR-R remained below 3.00 throughout. This pattern was observed across primary management, histology-driven pathways, platinum-resistant relapse, platinum-sensitive relapse, and complex reassessment scenes. The largest performance gaps between OMGs and CHAIR-R were observed in Evidence, Actionability, and Robustness, particularly in relapse and histology-driven pathway scenes.

Beyond mean SPEAR scores, we analysed the full Likert distribution to characterize safety boundaries and tail risk. Ratings were collapsed into three categories (low: 1–2, neutral: 3, high: 4–5) to quantify low-score prevalence across five clinical scenes (**Supplementary Fig. S1**). Full 5-point distributions were assessed to evaluate distributional shape and concentration across SPEAR dimensions (**Supplementary Fig. S2**). Unlike single-agent models, OMGs maintained Safety scores at or above 3 in all clinical scenes, reflecting built-in safeguards that suppress high-risk outputs. Instances with Safety scores of 3 corresponded consistently to predefined record-constrained failure modes: ambiguous disease status or histology, discordant clinical documentation, or unverified treatment constraints (**Supplementary Table S6**). These findings indicate that OMG safety is largely contingent on, rather than independent of, underlying clinical record quality.

**Phase II: Multicentre retrospective re-MDT evaluation**

We next conducted a multicentre retrospective re-MDT evaluation to assess the stability of OMGs across clinical scenes, institutions, and underlying base models, as well as its concordance with human MDT reassessment. A scene-balanced random subset of 100 cases from FUSCC was selected, with 20 cases sampled from each clinical scene, for detailed benchmarking across multiple large language models. We additionally performed external-centre validation on retrospective cohorts from NJPH and TFPH using the same Phase II protocol. All re-MDT assessments were conducted by a re-constituted multidisciplinary panel convened specifically for reassessment purposes. The panel followed the same five-role MDT structure (chair, medical oncology, radiology, pathology, and nuclear medicine) and included senior clinicians with routine cross-institutional practice who did not participate in the original MDT discussions for the included cases. Re-MDT conclusions were treated as an independent human reference arm rather than a definitive gold standard.

Across the FUSCC re-MDT cohort, OMGs powered by GPT-5.1 achieved the highest overall decision quality among evaluated backbones. In the scene-balanced cohort ($n = 100$; 20 per scene), the mean Safety-gated overall SPEAR score was $4.45 \pm 0.30$ for OMGs (GPT-5.1) and $4.53 \pm 0.23$ for re-MDT (**Supplementary Table S7**). Scene-stratified paired comparisons of the Safety-gated overall score showed no statistically significant differences between OMGs (GPT-5.1) and re-MDT after Bonferroni correction across the five scenes (paired two-sided Wilcoxon; **Fig. 3a**). At the dimension level, OMGs (GPT-5.1) achieved mean scores of $4.56 \pm 0.54$ (Safety), $4.26 \pm 0.56$ (Personalization), $4.57 \pm 0.54$ (Evidence), $4.18 \pm 0.86$ (Actionability), and $4.70 \pm 0.48$ (Robustness), compared with $4.74 \pm 0.44$, $4.76 \pm 0.43$,

3.92 ± 0.56, 4.88 ± 0.33, and 4.37 ± 0.51 for re-MDT, respectively (**Fig. 3b**). Paired tests across the five SPEAR dimensions confirmed a complementary performance profile: OMGs scored higher on Evidence and Robustness, whereas re-MDT scored higher on Actionability and Personalization (**Fig. 3b**). In contrast, other LLM backbones showed substantially lower mean performance and greater variability, with overall SPEAR scores ranging from 2.61 ± 0.40 to 4.14 ± 0.29 (**Supplementary Table S7–8**). Higher Evidence scores for OMGs reflect both enhanced traceability and more consistent evidence alignment: the system systematically retrieves guideline- and trial-level evidence and matches it to treatment line, biomarker status, and disease stage, which is difficult to perform exhaustively and document explicitly under routine, time-constrained MDT workflows.

Cross-centre analyses further demonstrated overall alignment (**Fig. 3c–e**). In the FUSCC cohort, OMGs (GPT-5.1) closely tracked re-MDT across all five SPEAR dimensions, with high concordance in Safety and Evidence. Similar patterns were observed in the NJPH cohort, where OMGs (GPT-5.1) achieved mean scores of 4.77 ± 0.57 for Safety, 4.50 ± 0.78 for Personalization, 4.77 ± 0.43 for Evidence, 4.67 ± 0.71 for Actionability, and 4.80 ± 0.41 for Robustness, compared with re-MDT scores of 4.97 ± 0.18, 4.93 ± 0.25, 4.40 ± 0.56, 5.00 ± 0.00, and 4.87 ± 0.35, respectively (**Supplementary Table S9**). In the TFPH cohort, OMGs (GPT-5.1) likewise demonstrated close agreement with re-MDT, achieving mean scores of 4.76 ± 0.44 for Safety, 4.67 ± 0.58 for Personalization, 4.90 ± 0.30 for Evidence, 4.62 ± 0.50 for Actionability, and 4.71 ± 0.46 for Robustness, compared with re-MDT scores of 4.81 ± 0.40, 4.67 ± 0.48, 4.48 ± 0.51, 4.90 ± 0.30, and 4.76 ± 0.44, respectively (**Supplementary Table S10**).

To contextualize Phase II feasibility, we reported per-case inference cost and runtime for OMGs, together with the frequency with which citation audits resulted in conservative Evidence-score capping. Under identical inference settings, the full OMGs framework required a median of

134,656 total tokens per case (IQR, 19,130) and achieved a median end-to-end latency of 155.3 s (IQR, 33.1), remaining within the time scales of scheduled multidisciplinary tumour board deliberation (**Supplementary Table S11**). Citation audits were performed solely to enforce predefined Evidence score capping rules. In the 100 scene-balanced audited cases, 98% did not trigger Evidence capping, while 2% were conservatively capped at 3 due to partially supported citations; no case was capped at ≤2 (**Supplementary Table S12**).

**Phase III: Prospective multicentre evaluation against real-world MDT decisions**

In the prospective multicentre evaluation, OMGs performance closely approximated routine MDT conclusions. Mean SPEAR differences (OMGs minus MDT) were small across all centres, with 95% confidence intervals remaining within the prespecified equivalence margin of ±0.5 points (**Fig. 4a and Supplementary Table S13**).

Scene-stratified analyses showed modest dimension-specific deviations. Evidence scores were higher for OMGs due to explicit source citation and systematic retrieval of relevant guidelines, trials, and patient-level evidence matched to case features. Time-constrained MDT documentation cannot consistently capture both dimensions. Actionability and Safety declined slightly in complex cases when inputs were ambiguous or conflicting. Most deviations remained within the equivalence margin, with no progressive negative shift across complexity levels (**Fig. 4b and Supplementary Table S13**).

When absolute performance was examined, OMGs achieved high mean SPEAR scores that closely approximated real MDT assessments across all participating centres (**Fig. 4c–e**). In the FUSCC prospective cohort, OMGs achieved mean scores of 4.51 for Safety, 4.46 for Personalization, 4.72 for Evidence, 4.31 for Actionability, and 4.67 for Robustness, compared with corresponding real MDT scores of 4.87, 4.64, 4.49, 4.79, and 4.69, respectively (**Supplementary Table S14**). Similar patterns were observed in external prospective cohorts.

At FOGH, OMGs achieved mean scores ranging from 4.30 to 4.80 across all dimensions, closely matching real MDT scores, including identical peak performance in Actionability. At SFMIH, OMGs reached or exceeded real MDT performance in Safety and Evidence, with mean scores of 5.00 in both dimensions, while maintaining comparable performance in Personalization, Actionability, and Robustness.

**Phase IV: Human–AI collaborative MDT decision-making**

We next evaluated OMGs within a human–AI collaborative workflow by having physicians rate the same patient cases under two conditions: human-only decision-making and OMGs-assisted decision-making. Analyses were stratified by institution, physician background, and clinical scene (**Fig. 5 and Supplementary Tables S15–16**). Across all panels, OMGs assistance increased the safety-gated overall score compared with human-only decisions (paired Wilcoxon signed-rank tests), indicating a consistent net improvement in end-to-end patient-level decision quality.

In the prospective FUSCC cohort, OMGs assistance improved resident physicians' ratings across all five SPEAR domains when examined at the scene-by-domain level (**Fig. 5a and Supplementary Table S15**). After panel-wise multiplicity control (Bonferroni correction across all scene×domain tests within the panel), improvements were robust across Scenes 1–3 for every domain (all adjusted $P \leq 4.8 \times 10^{-4}$), with the largest shifts observed for Evidence. Benefits persisted in later scenes but became more heterogeneous. In Scene 4, Robustness did not remain significant after correction, whereas Safety, Personalization, Evidence, and Actionability remained significant. In Scene 5, Actionability did not remain significant after correction, while the other domains remained significant.

A similar pattern was observed among physicians from non-tertiary hospitals within the same cohort (**Fig. 5b and Supplementary Table S16**). Scene-stratified analyses again showed consistent gains across domains in Scenes 1–3 (all adjusted P ≤ $8.9×10^{-4}$). In Scene 4, Actionability did not remain significant after correction despite improvements in the other domains. In Scene 5, gains were retained for Personalization, Evidence strength and traceability, and Actionability, whereas Safety and Robustness did not remain significant after correction, consistent with reduced power in this smaller stratum and greater dispersion in paired differences.

Generalizability was supported by external cohorts from FOGH and SFMIH, where OMGs assistance increased the safety-gated overall score relative to human-only decision-making (**Fig. 5c–d**). At the domain level, OMGs assistance yielded significant improvements in most domains in both cohorts (Supplementary Table S16), with the most consistent gains in Evidence strength and Robustness. Effect sizes and statistical stability varied by physician group, including attenuated Personalization gains among FOGH residents and attenuated Safety gains among SFMIH local physicians.

**Discussion**

In this study, we present OMGs, a multi-agent system that resembles real-world MDT workflows to support ovarian cancer decision-making. By coordinating five specialized agents, OMGs enables collaborative deliberation and generates traceable, evidence-backed recommendations. Unlike systems designed for outcome prediction or automated treatment selection, OMGs serves as a decision-support scaffold, making clinical assumptions, uncertainties, and evidence dependencies explicit at the point where synthesis matters most. Evaluated across four phases and five clinical scenes, OMGs consistently delivered high-quality decisions, aligned closely with MDT conclusions in both retrospective reassessments and prospective comparisons, and improved clinicians' written recommendations in human–AI collaborative workflows.

OMGs is motivated by persistent friction points in ovarian tumour MDT practice—time pressure, fragmented longitudinal documentation, limited traceability, and resource asymmetry across hospitals.[15,27] Recent multi-agent LLM systems, such as MDAgents and EvoMDT, have shown promise, but evaluations have primarily focused on benchmarks or standardized cases rather than routine MDT workflows.[12,23,25] For example, MDAgents was evaluated primarily on benchmark datasets;[25] EvoMDT uses five function-specific agents (diagnosis, treatment, safety, monitoring, and a coordinator) arranged by decisional step rather than by specialty role, and its evaluation relied on individual physician ratings of standardized case summaries rather than a convened multidisciplinary discussion;[23] the autonomous oncology agent study similarly relied on a small set of realistic but fictional multimodal oncology cases with blinded expert review.[12] In contrast, OMGs addresses both architectural and validation gaps: (i) each agent assumes a distinct specialty role, ensuring that the resulting deliberation reproduces the cross-disciplinary negotiation characteristic of actual tumour boards; (ii) we further validated the system on real-world ovarian tumour records across multiple centres through re-MDT reassessment, prospective comparison with routine MDT conclusions, and paired human–AI collaborative studies. The system also preserves the intermediate multi-role deliberation as a traceable dialogue that mirrors convened MDT discussions in its specialty workflow and information dependencies, capturing how viewpoints are raised, evidence-checked, and reconciled into a final summary (**Extended Data Fig. 1**). Together, this four-phase, multicentre evaluation provides a clinically grounded test of MDT-oriented multi-agent assistance under heterogeneous real-world records.

Two principal findings emerge from this work. The first concerns decision quality across clinical scenes. In the single-centre retrospective benchmark, OMGs outperformed the strongest single-agent baseline across all SPEAR dimensions, with the largest gains observed in Evidence and Robustness. This pattern reflects a clinical reality familiar to MDT practice: in complex ovarian tumour scenarios, the limiting factor is rarely the naming of a regimen, but rather the ability to justify a strategy, identify missing or conflicting variables, and define the conditions under which management should change. Performance remained stable when evaluated against a unified re-MDT reference standard in a scene-balanced, multicentre cohort, and prospective comparisons showed only modest deviations from routine MDT conclusions despite increasing clinical complexity. Taken together, these results suggest that OMGs can approximate real-world MDT reasoning while preserving documentation-driven interpretability under heterogeneous and imperfect records.

The second finding relates to near-term clinical utility. In paired human–AI evaluations, clinicians produced higher-quality MDT-style recommendations when supported by OMGs, with the most pronounced improvements observed among resident physicians and clinicians from non-tertiary hospitals. The gains were particularly evident in Evidence, Actionability, and Robustness, domains that are often underdeveloped when subspecialty input or formal MDT infrastructure is limited. In this context, OMGs appears to function as a structured cognitive scaffold, reducing omission risk and promoting explicit reasoning rather than replacing clinical judgment. The ability to externalize MDT-style deliberation into auditable, evidence-linked text may therefore help narrow practice variability and improve the defensibility of longitudinal decisions in resource-constrained settings.

OMGs provides practical value beyond algorithmic performance by reflecting real-world MDT workflows, where five specialized agents collaborate to reach a consensus. This collaborative approach generates transparent, traceable recommendations that can be reviewed and refined over time, ensuring that clinical decisions are based on a comprehensive, multi-disciplinary perspective.[28,29] With a median processing time of 155 seconds per case, OMGs pre-generates analyses before scheduled MDT meetings, seamlessly fitting into routine clinical workflows. For institutions with limited MDT resources, OMGs offers guideline-based decision support, maintaining institutional oversight while enabling collaborative decision-making. This is especially beneficial for smaller hospitals lacking dedicated MDT infrastructure, helping standardize evidence-based care across diverse clinical settings. Additionally, the system serves an educational role by documenting clinical reasoning, evidence sources, and reassessment criteria, making it an invaluable learning tool for trainees and junior physicians.

Despite these strengths, several limitations merit careful consideration. All evaluations were conducted within the Chinese healthcare system, and generalizability to settings with different languages, guideline ecosystems, formulary constraints, and MDT organization requires further investigation. While Phase IV suggests that the system may be especially useful under resource constraints, cross-regional validation will be necessary to assess how differences in practice patterns and evidence hierarchies affect performance. In addition, the evaluations were offline and non-interventional, focusing on decision quality and concordance rather than causal effects on patient outcomes. Accordingly, the present results should be interpreted as evidence of technical feasibility and potential clinical utility, rather than proof of improved survival or toxicity profiles. In Phase IV, improvements primarily reflect enhanced structured documentation, evidence articulation, and failure-mode surfacing under assistance; future randomized or counterbalanced designs will be required to disentangle assistance effects from carryover and anchoring. Finally, real-world EHR fragmentation and missingness remain a dominant source of residual risk. Although structured MDT-style outputs can surface missing or conflicting information, system performance and safety remain tightly coupled to input quality, underscoring the need for governance mechanisms, evidence-bank maintenance, and explicit escalation pathways in any future deployment.

In summary, OMGs demonstrates that a multi-agent system with role-specific constraints and traceable evidence can effectively replicate MDT-level decision-making for ovarian malignancies across both retrospective and prospective multicentre evaluations. The system consistently improves clinicians' recommendations in collaborative workflows. Instead of positioning LLMs as autonomous decision-makers, this work presents a model in which these systems enhance MDT deliberation by strengthening evidence integration, managing uncertainty, and improving documentation, while clinicians retain final decision-making responsibility. As health systems worldwide confront oncology workforce shortages and widening care disparities, such human–AI collaboration models may prove essential to sustaining high-quality cancer care at scale. Further validation across diverse health systems, coupled with prospective interventional studies and robust evidence governance, will be essential to fully assess the clinical and system-level impact of MDT-oriented LLM support.

**Main References**

**Tables**

**Table 1 | Baseline characteristics of retrospective cohorts**

| | ALL N=304 | FUSCC (retrospective) N=253 | NJPH N=30 | TFPH N=21 | P value |
|---|---|---|---|---|---|
| **Age** | 55.0 [47.0;62.0] | 55.0 [46.0;62.0] | 54.5 [49.5;60.8] | 52.0 [47.0;57.0] | 0.438 |
| **histology_group:** | | | | | 0.938 |
| EOC | 250 (82.2%) | 201 (79.4%) | 30 (100.0%) | 19 (90.5%) | |
| BET | 7 (2.3%) | 7 (2.8%) | 0 (0.0%) | 0 (0.0%) | |
| GCT | 17 (5.6%) | 16 (6.3%) | 0 (0.0%) | 1 (4.8%) | |
| SCST | 11 (3.6%) | 10 (4.0%) | 0 (0.0%) | 1 (4.8%) | |
| GCSCST | 1 (0.3%) | 1 (0.4%) | 0 (0.0%) | 0 (0.0%) | |
| NEN | 5 (1.6%) | 5 (2.0%) | 0 (0.0%) | 0 (0.0%) | |
| Sarcoma | 9 (3.0%) | 9 (3.6%) | 0 (0.0%) | 0 (0.0%) | |
| Other aggressive | 3 (1.0%) | 3 (1.2%) | 0 (0.0%) | 0 (0.0%) | |
| Unknown | 1 (0.3%) | 1 (0.4%) | 0 (0.0%) | 0 (0.0%) | |
| **FIGO 2014:** | | | | | 0.002 |
| I | 23 (7.6%) | 17 (6.7%) | 0 (0.0%) | 6 (28.6%) | |
| II | 20 (6.6%) | 16 (6.3%) | 2 (6.7%) | 2 (9.5%) | |
| III | 174 (57.2%) | 140 (55.3%) | 24 (80.0%) | 10 (47.6%) | |

| | | | | |
|---|---|---|---|---|
| IV | 87 (28.6%) | 80 (31.6%) | 4 (13.3%) | 3 (14.3%) |
| **Scene**: | | | | <0.001 |
| Scene1 | 64 (21.1%) | 54 (21.3%) | 2 (6.7%) | 8 (38.1%) |
| Scene2 | 49 (16.1%) | 47 (18.6%) | 0 (0.0%) | 2 (9.5%) |
| Scene3 | 72 (23.7%) | 61 (24.1%) | 3 (10.0%) | 8 (38.1%) |
| Scene4 | 41 (13.5%) | 33 (13.0%) | 8 (26.7%) | 0 (0.0%) |
| Scene5 | 78 (25.7%) | 58 (22.9%) | 17 (56.7%) | 3 (14.3%) |
| **Primary Treatment Strategy**: | | | | 0.286 |
| PDS | 245 (80.6%) | 198 (78.3%) | 27 (90.0%) | 20 (95.2%) |
| NACT+IDS | 54 (17.8%) | 50 (19.8%) | 3 (10.0%) | 1 (4.8%) |
| PST | 5 (1.6%) | 5 (2.0%) | 0 (0.0%) | 0 (0.0%) |

Abbreviations: FUSCC, Fudan University Shanghai Cancer Center; NJPH, Northern Jiangsu People's Hospital; TFPH, Taizhou First People's Hospital; EOC, epithelial ovarian carcinoma; BET, borderline epithelial tumour; GCT, germ cell tumour; SCST, sex cord–stromal tumour; GCSCST, germ cell–sex cord-stromal tumour; NEN, neuroendocrine neoplasm; PDS, primary debulking surgery; NACT+IDS, neoadjuvant chemotherapy plus interval debulking surgery; PST, primary systemic therapy.

**Figure legends**

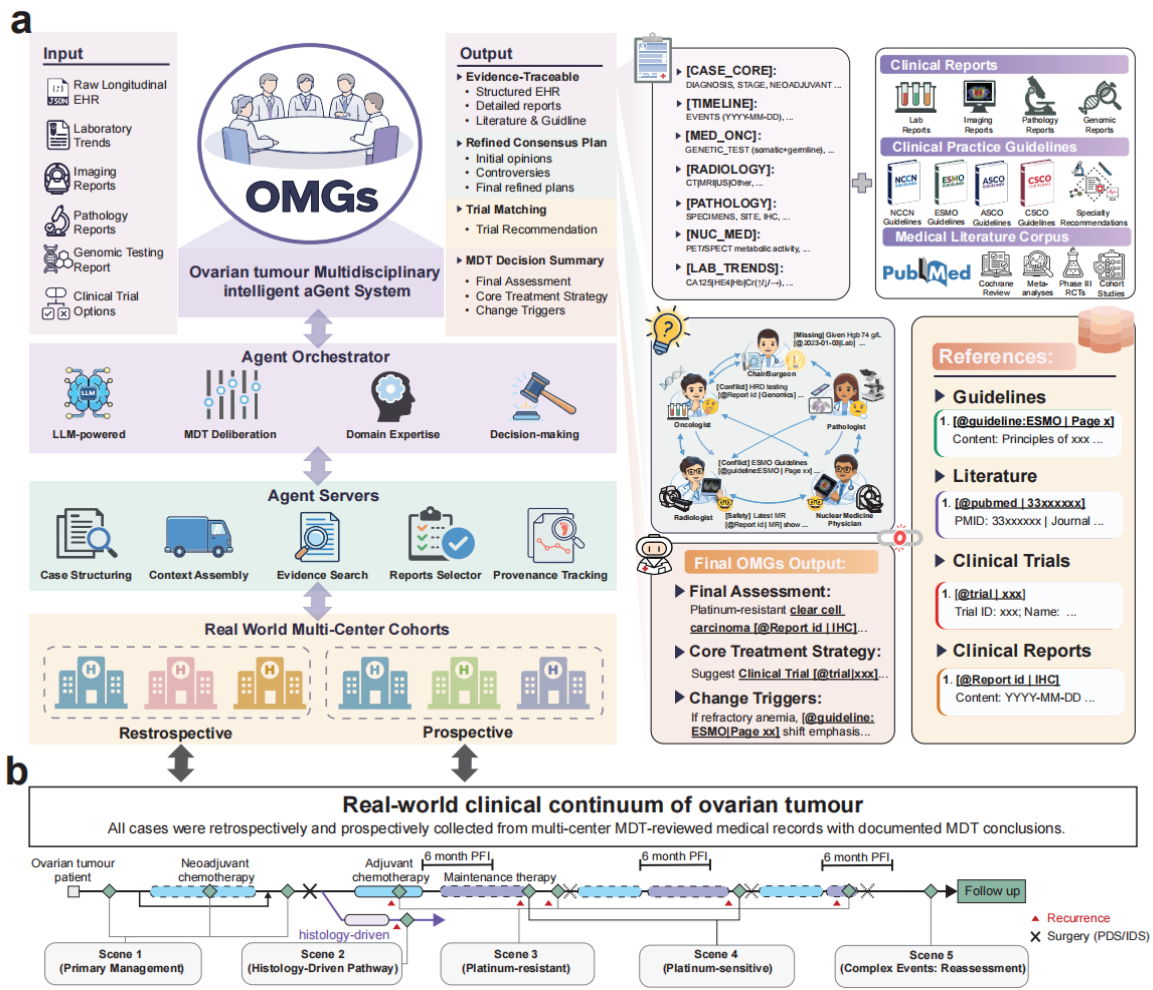

**Fig. 1 | Overview of OMGs and the five MDT clinical scenes across the ovarian tumour care continuum.**

**a**, OMGs workflow showing longitudinal EHR inputs, evidence retrieval and the structured MDT-style output with traceable references. **b**, Five prespecified MDT clinical scenes across the real-world ovarian tumour care continuum.

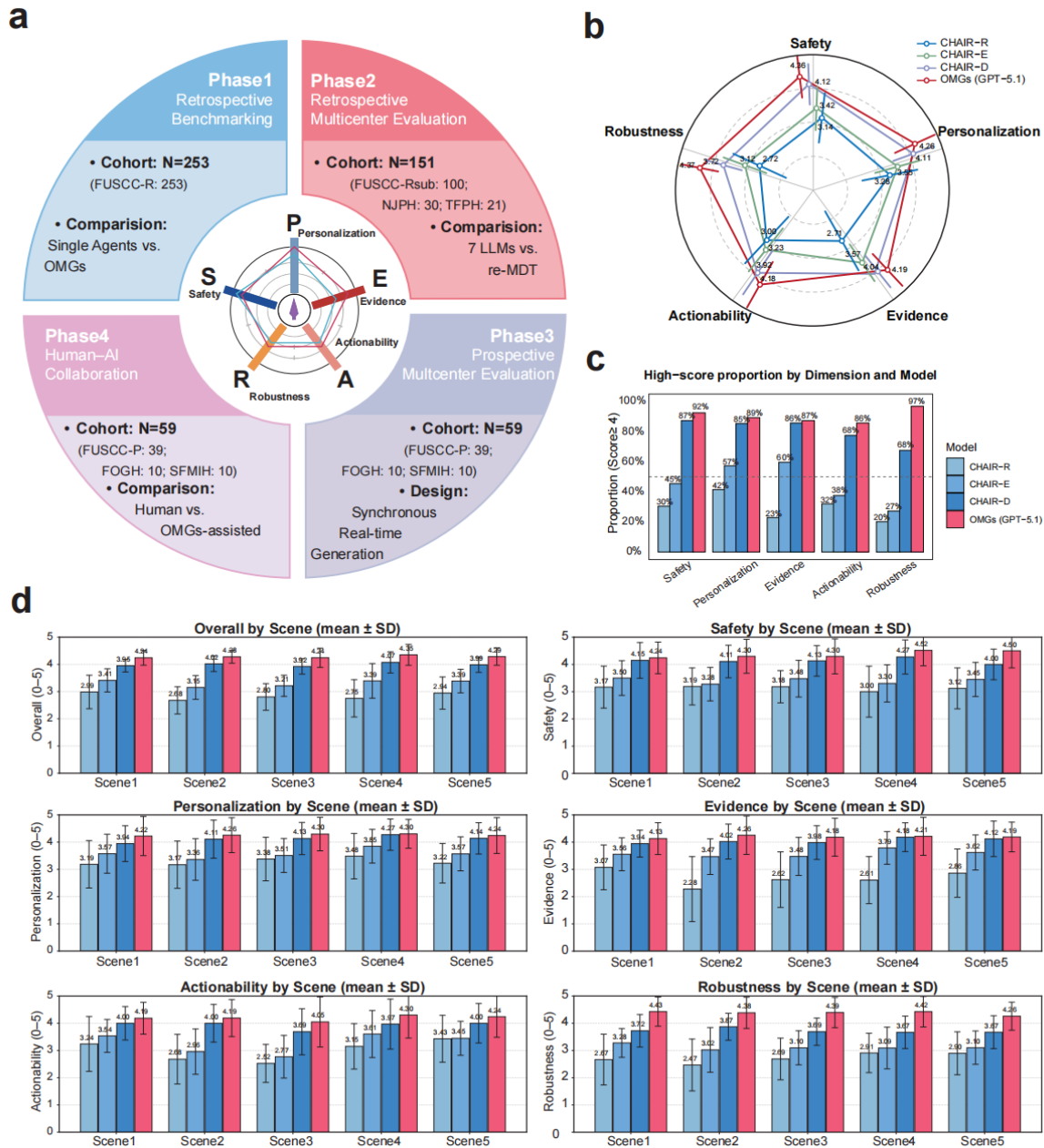

**Fig. 2 | Phase I single-centre retrospective benchmarking of OMGs using the SPEAR rubric.**

**a**, Four-phase study framework and the five SPEAR dimensions (Likert 1–5), with cohort sizes and comparison designs. **b**, Radar plot of mean SPEAR dimension scores across 253 retrospective cases from FUSCC comparing OMGs (GPT-5.1) with three CHAIR baselines (CHAIR-R, CHAIR-E and CHAIR-D). **c**, Proportion of high scores (≥4) by dimension and model. d, Scene-stratified mean (± s.d.) overall and dimension scores across the five predefined MDT clinical scenes.

Abbreviations: FUSCC, Fudan University Shanghai Cancer Center; s.d., standard deviation.

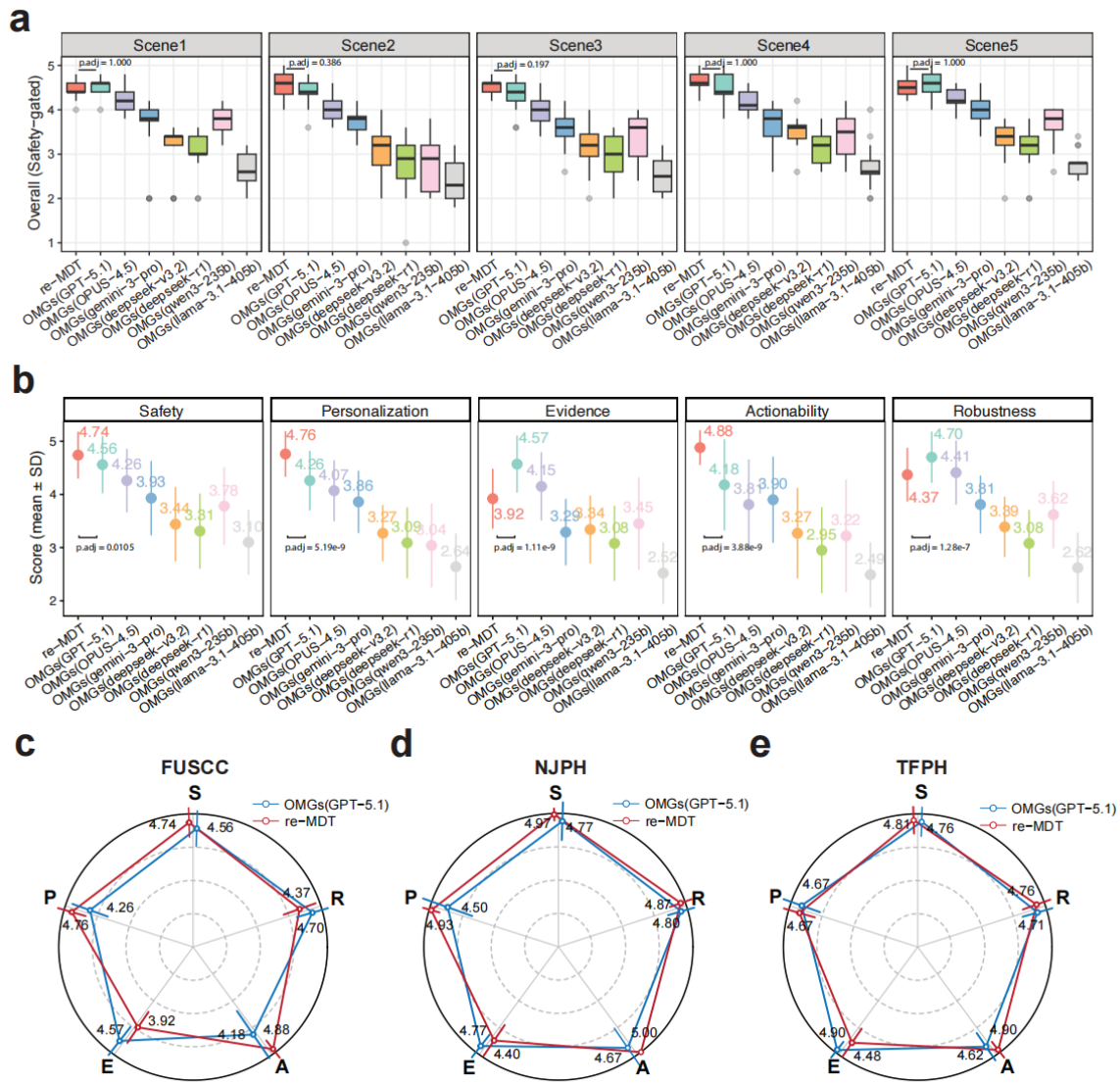

**Fig. 3 | Phase II multicentre retrospective re-MDT study benchmarking OMGs against re-MDT across clinical scenes and centres.**

a, Scene-stratified distributions of the Safety-gated overall score for the re-MDT human reassessment arm and OMGs instantiated with different backbone models in the scene-balanced FUSCC cohort (n = 100, 20 per scene). Boxes show the median and interquartile range. *P* values are from paired two-sided Wilcoxon tests comparing re-MDT and OMGs (GPT-5.1) within each scene (paired by case), with Bonferroni correction across the five scenes. **b**, Mean ± s.d. SPEAR dimension scores for the same cohort, comparing re-MDT with each OMGs backbone. *P* values are from paired two-sided Wilcoxon tests comparing re-MDT and OMGs (GPT-5.1) for each SPEAR dimension (paired by case), with Bonferroni correction across the five dimensions. **c–e**, Centre-level radar plots of mean SPEAR profiles comparing OMGs (GPT-5.1) with re-MDT across the FUSCC retrospective cohort (**c**), NJPH cohort (**d**) and TFPH cohort (**e**).

Abbreviations: re-MDT, re-constituted multidisciplinary tumour board reassessment; FUSCC, Fudan University Shanghai Cancer Center; NJPH, Northern Jiangsu People's Hospital; TFPH, Taizhou First People's Hospital; s.d., standard deviation.

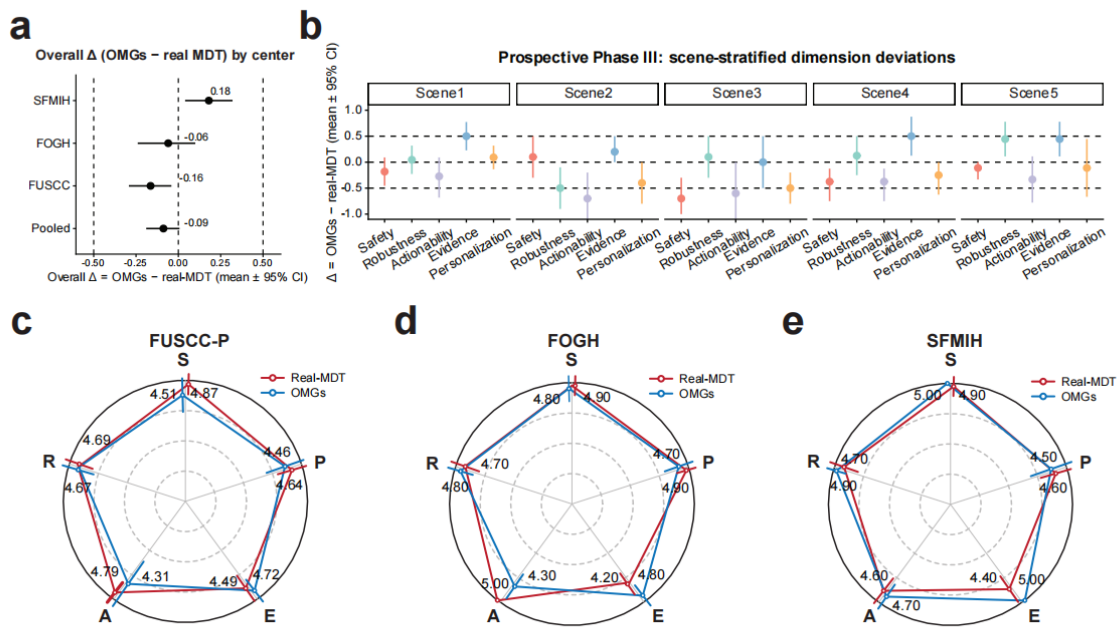

**Fig. 4 | Phase III prospective multicentre evaluation of OMGs against real-world MDT decisions.**

**a**, Centre-level mean difference in Safety-gated overall score between OMGs and routine MDT decisions (Δ = OMGs − real MDT), shown as mean ± 95% confidence interval (CI) for each centre and the pooled cohort. Vertical dashed lines indicate the prespecified equivalence margin (±0.5 points). **b**, Scene-stratified dimension-level deviations (Δ = OMGs − real MDT) across the five MDT clinical scenes. Points show mean Δ with 95% CI for each SPEAR dimension; horizontal dashed lines mark 0 and the equivalence bounds (±0.5). **c–e**, Radar plots of absolute mean SPEAR dimension scores comparing OMGs with routine MDT decisions in the prospective cohorts from FUSCC (**c**), FOGH (**d**) and SFMIH (**e**).

Abbreviations: FUSCC, Fudan University Shanghai Cancer Center; FOGH, Fudan University Obstetrics and Gynecology Hospital; SFMIH, Shanghai First Maternity and Infant Hospital; CI, confidence interval.

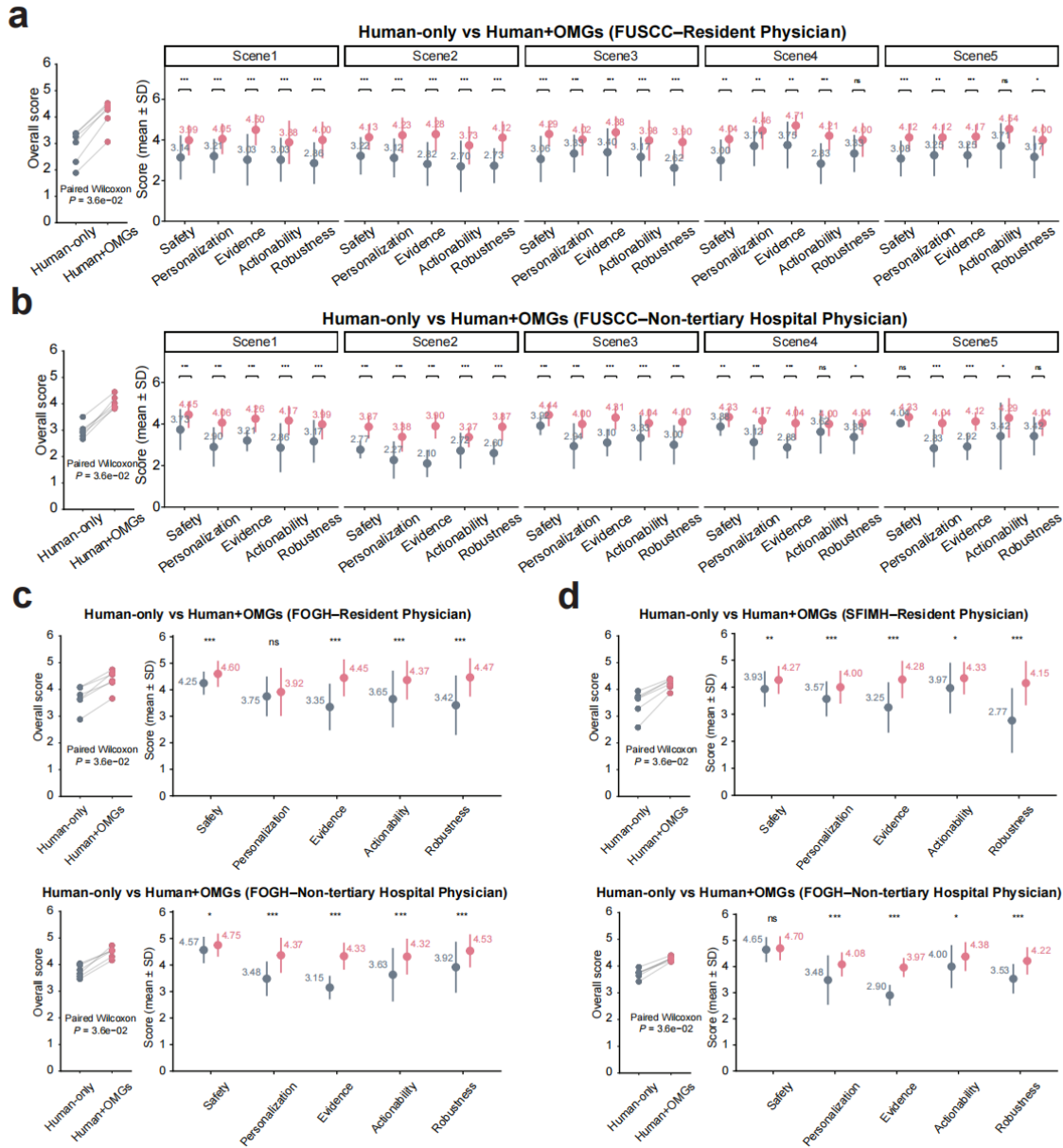

**Fig. 5 | Phase IV human–AI collaborative MDT decision-making with OMGs assistance.**

**a**, Paired comparison between human-only and OMGs-assisted decision-making in the FUSCC prospective cohort for resident physicians, shown as paired changes in Safety-gated overall score (left) and scene-stratified mean ± s.d. SPEAR dimension scores (right). **b**, Same analysis in the FUSCC prospective cohort for non-tertiary hospital physicians. **c**, External prospective validation in the FOGH cohort, shown separately for resident physicians (top) and non-tertiary hospital physicians (bottom), with paired overall-score shifts (left) and mean ± s.d. SPEAR dimension scores (right). **d**, External prospective validation in the SFMIH cohort, shown separately for resident physicians (top) and non-tertiary hospital physicians (bottom), with paired overall-score shifts (left) and mean ± s.d. SPEAR dimension scores (right). Across panels, $P$ values are from paired Wilcoxon signed-rank tests; dimension-level significance annotations reflect Bonferroni correction within each panel across all scene × dimension comparisons.

Abbreviations: FUSCC, Fudan University Shanghai Cancer Center; FOGH, Fudan University Obstetrics and Gynecology Hospital; SFMIH, Shanghai First Maternity and Infant Hospital; s.d., standard deviation.

# Methods

## Ethics approval and study oversight

Multicentre electronic health record (EHR) and clinical cohort data were derived from tertiary referral hospitals in Shanghai, Jiangsu Province, and Zhejiang Province, including Fudan University Shanghai Cancer Center (FUSCC), Northern Jiangsu People's Hospital (NJPH), Taizhou First People's Hospital (TFPH), Fudan University Obstetrics and Gynecology Hospital (FOGH), and Shanghai First Maternity and Infant Hospital (SFMIH). The dataset comprised both retrospective and prospective cohorts, incorporating structured and semi-structured data on demographics, pathology, imaging, laboratory tests, molecular and genetic profiling, treatment strategies, longitudinal outcomes, and follow-up records, thereby enabling large-scale evaluation of clinical decision-support performance in real-world oncology practice. Ethics approval was obtained from the institutional review boards of all participating centres. Retrospective analyses were conducted under an ethics approval obtained at FUSCC (protocol 2601-Exp365) using fully de-identified data. Prospective cohort studies were approved at FUSCC (protocol 2508-Exp538), Shanghai First Maternity and Infant Hospital (acceptance number KS25394), and the Obstetrics and Gynecology Hospital of Fudan University (protocol 2025-152). All study procedures were conducted in accordance with the principles of the Declaration of Helsinki and applicable national regulations for biomedical research.

For retrospective analyses, a waiver of informed consent was granted because the study involved minimal risk and used only fully de-identified data. For prospective cohorts, written informed consent was obtained prior to enrollment. The OMGs system was evaluated in a strictly non-interventional and observational manner; its outputs were generated offline and were not disclosed to clinicians or used in clinical decision-making. All OMGs runs used de-identified case packets prepared for research evaluation; no direct patient identifiers were provided to the system.

**Clinical input processing**

For each case, OMGs converts longitudinal EHR documents into a structured case summary aligned to the index MDT date. The index MDT date was defined as the recorded date of MDT discussion. Given a raw case record $X^{raw}$, comprising heterogeneous clinical documents and associated metadata, the system applies an EHR structuring function $f_{struct}$ to generate a normalized case representation $X$:

$$X = f_{struct}(X^{raw})$$

Using a prespecified case template, an LLM extracts explicit facts from each document and outputs a single schema-conformant JSON (missing or unsupported values recorded as "Unknown"). Extracted items retain document-level provenance metadata (e.g., document type and document date) to support traceability of the structured fields to source materials. Extracted items are standardized and merged across documents using modality- and time-based rules, without imputing unavailable information (**Supplementary Table S17**).

**Role-scoped information access and evidence retrieval**

To reflect real-world MDT boundaries, we enforce role-scoped access by packaging specialty-specific inputs from the structured case representation and prespecified source documents. Specialist agents are restricted to role-relevant inputs, whereas the chair agent has access to its own role-scoped package and integrates cross-specialty contributions during deliberation, with final arbitration and synthesis at the end (**Supplementary Table S18–19**).

External medical knowledge is incorporated through a controlled evidence retrieval module driven by queries generated from the structured case schema, with query construction anchored to disease context, clinical scene, and prior treatment exposures. As illustrated in **Fig. 1a**, retrieved evidence is organized into two complementary streams retrieved from two pre-chunked, semantically embedded corpora using embedding-similarity search. First, clinical practice guidelines are curated and harmonized across major international and national frameworks, and augmented with specialty-specific recommendations required to operationalize guideline-concordant ovarian tumour MDT decisions. Second, medical literature is drawn from publicly released MEDLINE baseline dataset, prioritizing higher-tier clinical evidence, including Cochrane reviews, meta-analyses, phase III randomized controlled trials, and well-designed cohort studies. Retrieved items are deduplicated, filtered, and normalized into stable evidence-bank entries; downstream citations are restricted to evidence-bank entries or verifiable patient-report provenance. Evidence-bank entries were assigned stable identifiers and stored with bibliographic fields (PMID when available) to support auditable citation linkage. To avoid temporal drift during evaluation, the guideline corpus and evidence-bank index used for retrieval were fixed for each prespecified evaluation phase.

**Multi-agent deliberation and schema-constrained output**

Conditioned on the structured case representation, role-scoped inputs, and a centralized evidence bank, the system conducts controlled multi-role deliberation that mirrors MDT governance rather than free-form agent interaction. Each role agent first produces an independent, role-specific assessment and recommendation, explicitly stating safety considerations, key uncertainties, and traceable supporting references restricted to evidence-bank entries (**Supplementary Fig. S1**).

Following this initial phase, a constrained deliberation stage is triggered in which agents intervene selectively rather than continuously. By default, agents remain silent and are permitted to contribute only when predefined conditions are met, including inter-role conflicts, safety concerns, missing critical information, or newly identified decision-relevant evidence. All interventions are role-directed and explicitly annotated with their rationale, ensuring focused and auditable exchanges while preventing diffuse or redundant discussion.

The chair agent subsequently reconciles cross-specialty disagreements and synthesizes a unified MDT-style recommendation expressed in a schema-constrained form:

$$Y = (Final\ Assessment, Core\ Treatment\ Strategy, Change\ Triggers)$$

where Final Assessment summarizes the reconciled disease status and risk stratification, Core Treatment Strategy specifies the selected management pathway and its clinical rationale, and Change Triggers define explicit clinical or safety conditions under which reassessment or escalation is warranted. Each element of $Y$ is required to be accompanied by traceable evidence references to enable auditable evaluation of decision quality.

**Cohorts, eligibility, and patient selection**

We assembled retrospective and prospective cohorts across five hospitals, using prespecified eligibility criteria designed to ensure that each case could support reconstruction of key clinical variables and an auditable MDT-style evaluation. Eligible patients were required to have (i) a confirmed ovarian tumour diagnosis and sufficient documentation to support ascertainment of core clinicopathologic features as available at the time of MDT discussion, with explicit designation of unknown when definitive classification was not possible, (ii) availability of specialty-relevant source materials (pathology, imaging, and molecular or laboratory data when applicable), and (iii) MDT-related documentation adequate for downstream evaluation.

Patients with incomplete clinical information or incomplete MDT documentation were excluded, consistent with the study flow described in the **Supplementary Fig. S2**. For retrospective analyses, de-identified real-world clinical records were screened across FUSCC, Northern Jiangsu People's Hospital (NJPH), and Taizhou First People's Hospital (TFPH). A total of 304 retrospective cases were included (FUSCC, n = 253; NJPH, n = 30; TFPH, n = 21). For prospective evaluation, consecutive patients undergoing routine MDT discussion were enrolled at FUSCC, Fudan University Obstetrics and Gynecology Hospital (FOGH), and Shanghai First Maternity and Infant Hospital (SFMIH), with additional requirements of written informed consent and contemporaneous MDT documentation. A total of 59 prospective cases were included (FUSCC, n = 39; FOGH, n = 10; SFMIH, n = 10).

**Clinical scenes, comparators, and evaluation design**

To standardize case stratification across institutions and study phases, all eligible patients were assigned a priori to one of five prespecified MDT clinical scenes (**Fig. 1b**): primary management, histology-driven pathways, platinum-resistant relapse, platinum-sensitive relapse, and complex events requiring reassessment. Scene assignment rules were finalized before evaluation and applied uniformly across centres and phases (**Supplementary Table S1**).

The evaluation comprised four sequential phases designed to progressively increase external validity and workflow realism (**Fig. 2a and Supplementary Fig. S3).** Phase I benchmarked OMGs against single-agent baselines under blinded evaluation. Original MDT conclusions were provided as contextual reference to support Actionability scoring. Raters were instructed to use original MDT conclusions only for Actionability assessment and not as a substitute anchor for Safety or Evidence judgments. Phase II was a retrospective re-evaluation using a scene-balanced subset of 100 cases from FUSCC (20 per scene), with external multicentre validation in independent cohorts from NJPH ($n = 30$) and TFPH ($n = 21$). The re-MDT comprised five roles (chair, medical oncology, radiology, pathology, and nuclear medicine); none of the participating clinicians had been involved in the original MDT discussions. Re-discussion used the standardized case packet restricted to information available at the time of original decision. Re-MDT conclusions were treated as an independent human comparator rather than a definitive gold standard. Phase III consisted of a prospective, multicentre comparison in which OMGs outputs were evaluated against contemporaneous MDT conclusions generated during routine clinical care. Phase IV evaluated human–AI collaboration by comparing physician decision-making with and without OMGs assistance across different hospital settings.

Across phases, prespecified procedures were implemented to reduce bias and preserve comparability. Where applicable, evaluators were blinded to comparator identity, and the re-MDT team in Phase II was blinded to original MDT conclusions and system outputs. In Phase I, three senior gynaecologic oncology experts independently rated system outputs under blinded conditions, with original MDT conclusions provided as contextual reference for Actionability scoring. Median scores were used and inter-rater agreement was assessed. In Phase II, SPEAR scoring was performed by an independent expert consensus panel that did not attend the re-MDT and was blinded to comparator identity and model assignment. The same consensus-based evaluation framework was applied in Phases III and IV, where MDT conclusions served as comparators rather than fixed gold standards.

In Phase I, we evaluated three single-agent chair comparators with progressively enriched information contexts. CHAIR-R used the schema-normalized case representation only. CHAIR-E added retrieval-augmented generation while keeping the same structured case inputs. CHAIR-D further added a preassembled, case-specific evidence dossier with stable report identifiers and matched trial eligibility information. For benchmarking, CHAIR-D and the full OMGs framework were input-matched at the index decision timepoint, including the same case representation, the same RAG pipeline, the same dossier sources and identifiers, and the same trial eligibility inputs. All outputs were generated using the same schema-constrained MDT template with identical section headers and length limits and were evaluated under blinded identity. Thus, the incremental effect estimate isolates the contribution of role-scoped deliberation and chair arbitration under matched inputs. CHAIR-D produces a single-agent chair synthesis, whereas OMGs performs role-scoped multi-agent deliberation followed by chair arbitration. Clinical inputs, evidence access, tooling, and output formatting were held constant across these two conditions, as summarized in **Supplementary Table S20**. The Phase II re-MDT was designed as a controlled human reference arm rather than an institution-independent ground truth.

**SPEAR rubric and safety-gated overall scoring**

System outputs were evaluated using the prespecified five-dimension Likert rubric (SPEAR): Safety (S), Personalization (P), Evidence strength and traceability (E), Actionability (A), and Robustness (R) (**Supplementary Table S2**). Each dimension was rated on a 5-point scale with anchors locked before evaluation. In Phase I, three gynaecologic oncology specialists with routine MDT practice scored each case independently, and the median score was used for each dimension. From Phase II onward, to avoid self-scoring when comparing OMGs with human MDT conclusions, SPEAR scoring was performed by an independent senior expert consensus panel that did not participate in the MDT or re-MDT discussions. The panel comprised five senior experts (gynaecologic oncology, medical oncology, radiology, pathology, and nuclear medicine) from tertiary referral hospitals with routine multidisciplinary tumour board practice and sustained ovarian cancer caseload. Raters were calibrated using representative cases spanning all five scenes. Disagreements were resolved using prespecified adjudication, with consensus discussion and escalation to a senior adjudicator when required.

Overall decision quality was computed as the mean across the five dimensions:

$$Overall_{raw} = \frac{S + P + E + A + R}{5}$$

To enforce safety as a hard constraint and prevent high aggregate scores from masking unsafe recommendations, we applied a safety-gated overall score:

$$Overall = \begin{cases} \min(Overall_{raw}, S), S < 3 \\ Overall_{raw}, S \geq 3 \end{cases}$$

In the human–AI collaboration phase, participating physicians first generated MDT-style recommendations under human-only conditions and then reassessed the same cases with access to OMGs outputs and evidence artifacts. This paired design ensured that each clinician served as their own control for each case. Analyses were prespecified to be stratified by institution, clinician background, and clinical scene. To minimize stylistic cues and length-driven bias, all recommendations were normalized into the same schema-constrained MDT template with fixed section headers and length limits. This normalization did not introduce any additional clinical content or supporting sources beyond the original case packet. Recommendations were rated under blinded comparator identity. Human MDT and re-MDT outputs were asked to provide identifiable supporting guideline or trial sources when making explicit evidence-based claims, enabling traceability scoring for the Evidence dimension.

**Citation fidelity audit and Evidence capping**

To prevent inflated Evidence ratings driven by superficial citation patterns, we implemented a conservative citation fidelity audit with rule-based Evidence capping. Evidence scores were first assigned according to the prespecified Likert anchors and then capped based on audit outcomes, ensuring that higher Evidence ratings reflect verifiable citation support rather than citation quantity. Citation traceability was operationalized by auditing only the citations explicitly invoked in the final recommendation for each case. After assigning the initial Evidence score using the rubric anchors (**Supplementary Table S2**), we applied predefined caps based on citation fidelity. If any invoked citation was partially supported, the Evidence score was capped at 3 or below. If any invoked citation was unsupported or hallucinated, the

Evidence score was capped at 2 or below (**Supplementary Table S13**). Citation support was assessed at the claim level: a citation was considered "supported" only if the cited source explicitly substantiated the specific clinical claim as written; otherwise it was judged partially supported or unsupported. The audit acted as an upper-bound constraint on Evidence scoring rather than a re-scoring mechanism.

Citation audits were conducted in Phase II on a scene-balanced subset sampled from the Phase I retrospective cohort. Each invoked citation was independently reviewed by two gynaecologic oncology clinicians, with discrepancies resolved by a third senior adjudicator. This audit provided an explicit, rule-based constraint on Evidence scoring grounded in citation fidelity.

**Human–AI collaborative workflow**

Phase IV assessed how clinicians' recommendations change when OMGs is available during case write-up. Twelve physicians participated, including six resident physicians with routine MDT exposure at tertiary referral hospitals and six physicians from non-tertiary hospitals. This prespecified grouping was used for stratified analyses.

Physicians completed two passes on the same cases. In the human-only pass, they reviewed standardized case materials and produced an MDT-style recommendation that included assessment, treatment plan, and explicit triggers for reassessment. They could consult routine references such as clinical guidelines and PubMed-indexed literature, but were instructed not to use any generative AI tools. In the assisted pass, physicians revisited the same cases with access to the complete OMGs output package, including the structured case summary and evidence artifacts, and could reuse, revise, or replace the system text to produce a final recommendation. All outputs were generated offline for research purposes and were not used in clinical decision-making. The assisted pass did not introduce additional patient information beyond the standardized case materials; it provided only OMGs-generated synthesis text and linked evidence identifiers.

Both human-only and assisted responses were evaluated using the same SPEAR rubric and scoring procedures as in system evaluation. Comparisons were performed within physician–case pairs, with analyses stratified by institution, physician group, and clinical scene.

**Statistical analysis**

All analyses were prespecified and performed at the case level unless otherwise stated. Continuous variables are summarized as mean ± standard deviation (s.d.) or median (interquartile range, IQR), and categorical variables as counts and percentages. The primary endpoints were the five SPEAR dimension scores and the safety-gated overall score, with a two-sided α of 0.05.

In Phase II, inter-rater agreement among three senior gynaecologic oncology experts was assessed using two-way random-effects intraclass correlation coefficients for average measures ($ICC(2,k)$), quantifying absolute agreement of aggregated expert judgments across the five SPEAR dimensions. Median expert scores were used for downstream analyses. Differences in each SPEAR dimension and the safety-gated overall score were tested with paired two-sided Wilcoxon signed-rank tests.

Phase III was designed to assess concordance between OMGs and routine MDT conclusions using a paired-case framework. An a priori equivalence-based sample size calculation (Paired T-Tests for Equivalence; PASS 2023 v23.0.2) assumed an equivalence margin of ±0.5 points for overall SPEAR, a conservative s.d. of paired differences of 1, overall α = 0.05 (two one-sided α = 0.025) and 90% power, yielding 54 evaluable paired cases. We prospectively evaluated 59 consecutive paired cases.

Phase IV evaluated human–AI collaboration using a paired physician–case design, in which each physician completed both human-only and OMGs-assisted recommendations for the same cases. Patient-level overall performance within each institution, physician group, and clinical scene was summarised at the physician level by averaging the safety-gated overall score across cases, and human-only versus assisted performance was compared using paired Wilcoxon signed-rank tests (paired by physician). For scene-by-domain analyses, differences between conditions were estimated using linear mixed-effects models with condition as a fixed effect and random intercepts for case and physician $score \sim condition + (1|case) + (1|physician)$. $P$ values for the mixed-effects models were obtained using standard software implementations, and multiple comparisons were controlled using Bonferroni adjustment.

Categorical outcomes were compared using $\chi^2$ tests or Fisher's exact tests as appropriate; when exact tests were computationally infeasible for sparse contingency tables, Monte Carlo simulation was used to estimate $P$ values; effect sizes were reported as odds ratios or risk differences and 95% confidence intervals; for analyses involving multiple independent comparators, multiplicity was controlled using the Benjamini–Hochberg procedure ($q<0.05$).

**Data availability**

The study protocol is provided in the Supplementary Information. Source data are available via Github at https://github.com/pigudog/OMGs. Raw data are not publicly available due to the need to protect participant privacy, in accordance with the ethical approval for this study. Anonymized, non-dialogue individual-level data underlying the results can be requested by qualified researchers for academic use. Requests should include a research proposal, statistical analysis plan and justification for data use and can be submitted via email to X.W. (wu.xh@fudan.edu.cn). All requests will be reviewed by the Fudan University Shanghai

Cancer Center, and approved requests will be granted access via a secure platform after execution of a data access agreement.

**Code availability**

Comparative statistical analyses are detailed in the Article. Code for Core modules and prompts for OMGs is available via GitHub at https://github.com/pigudog/OMGs.

**Acknowledgements**

We would like to thank all doctors, nurses, patients, and their family members for their kindness in supporting our study. We also thank Dr. Zezhou Wang for his valuable guidance on statistical analysis. This work was supported by the National Natural Science Foundation of China (Nos. 82272898, 82203723, 82271940 and 82471932), the Three-Year Action Plan to Promote Clinical Skills and Clinical Innovation Capacity of Municipal Hospitals by the Shanghai Shenkang Hospital Development Center (SHDC2020CR5003-001), the Science and Technology Commission of Shanghai Municipality (21ZR1415000), and the Scientific Research Project of Taizhou Science and Technology Bureau in Zhejiang Province (23ywb47).

**Author contributions**

Y.Y.Z., Z.L.W., and S.W. conceived the study and designed the methodology. S.W. contributed significantly to the development of the methodology, including the AI framework and algorithm. C.H. supervised the study and provided guidance on the methodology. D.S.L. contributed to the design and development of the methodology. J.B.X., Y.Q.C., and Q.H.G. performed statistical analysis and interpreted the experimental results. Z.H.Z., S.L., J.L., J.D., H.L., J.Z., and Y.P.Z. performed clinical data collection and analysis. H.P.Z., W.J.W., H.Y.Z., and W.J. provided experimental materials, clinical data, and relevant resources for the study. Z.Q.L., Z.F., H.W. X.Z.J, and J.X.C. contributed to data processing, annotation, and organization. Y.Y.Z. and Z.L.W. wrote the initial draft of the manuscript. C.H., X.H.W., S.W., H.M.L., and Q.H.G. reviewed and edited the manuscript. Z.L.W. and S.W. contributed to visualizing the results and developing figures. C.H., S.W., and X.H.W. supervised the research and coordinated the collaborative efforts. H.M.L., X.H.W. and Q.H.G. managed the project, including coordinating resources and overseeing the progress. H.P.Z., H.M.L., X.H.W. and Q.H.G. provided financial support for the study and acquisition of research funding.

**Competing interest**

The authors declare no competing interests.

# Extended Data Tables

**Extended Data Table 1 | Baseline characteristics of prospective cohorts**

|  | ALL N=59 | FUSCC (prospective) N=39 | FOGH N=10 | SFMIH N=10 | P value |
|---|---|---|---|---|---|
| **Age** | 56.0 [50.0;65.0] | 55.0 [46.5;63.0] | 55.5 [54.0;57.5] | 62.0 [58.0;68.0] | 0.104 |
| **histology_group:** |  |  |  |  | 0.408 |
| EOC | 49 (83.1%) | 31 (79.5%) | 10 (100.0%) | 8 (80.0%) |  |
| GCT | 2 (3.4%) | 2 (5.1%) | 0 (0.0%) | 0 (0.0%) |  |
| SCST | 3 (5.1%) | 3 (7.7%) | 0 (0.0%) | 0 (0.0%) |  |
| Sarcoma | 2 (3.4%) | 2 (5.1%) | 0 (0.0%) | 0 (0.0%) |  |
| Other aggressive | 1 (1.7%) | 1 (2.6%) | 0 (0.0%) | 0 (0.0%) |  |
| Unknown | 2 (3.4%) | 0 (0.0%) | 0 (0.0%) | 2 (20.0%) |  |
| **FIGO_group:** |  |  |  |  | 0.001 |
| I | 1 (1.7%) | 1 (2.6%) | 0 (0.0%) | 0 (0.0%) |  |
| II | 5 (8.5%) | 3 (7.7%) | 1 (10.0%) | 1 (10.0%) |  |
| III | 26 (44.1%) | 13 (33.3%) | 8 (80.0%) | 5 (50.0%) |  |
| IV | 24 (40.7%) | 22 (56.4%) | 1 (10.0%) | 1 (10.0%) |  |

| | | | | |
|---|---|---|---|---|
| Unknown | 3 (5.1%) | 0 (0.0%) | 0 (0.0%) | 3 (30.0%) | |
| **Scene:** | | | | | 0.009 |
| Scene1 | 22 (37.3%) | 13 (33.3%) | 2 (20.0%) | 7 (70.0%) | |
| Scene2 | 10 (16.9%) | 10 (25.6%) | 0 (0.0%) | 0 (0.0%) | |
| Scene3 | 10 (16.9%) | 8 (20.5%) | 2 (20.0%) | 0 (0.0%) | |
| Scene4 | 8 (13.6%) | 4 (10.3%) | 4 (40.0%) | 0 (0.0%) | |
| Scene5 | 9 (15.3%) | 4 (10.3%) | 2 (20.0%) | 3 (30.0%) | |
| **Primary Treatment Strategy:** | | | | | 0.465 |
| PDS | 44 (74.6%) | 27 (69.2%) | 9 (90.0%) | 8 (80.0%) | |
| NACT+IDS | 15 (25.4%) | 12 (30.8%) | 1 (10.0%) | 2 (20.0%) | |

Abbreviations: FUSCC, Fudan University Shanghai Cancer Center; FOGH, Obstetrics and Gynecology Hospital of Fudan University; SFMIH, Shanghai First Maternity and Infant Hospital; EOC, epithelial ovarian carcinoma; GCT, germ cell tumour; SCST, sex cord–stromal tumour; FIGO, International Federation of Gynecology and Obstetrics; PDS, primary debulking surgery; NACT+IDS, neoadjuvant chemotherapy plus interval debulking surgery.

# Extended Data Figures

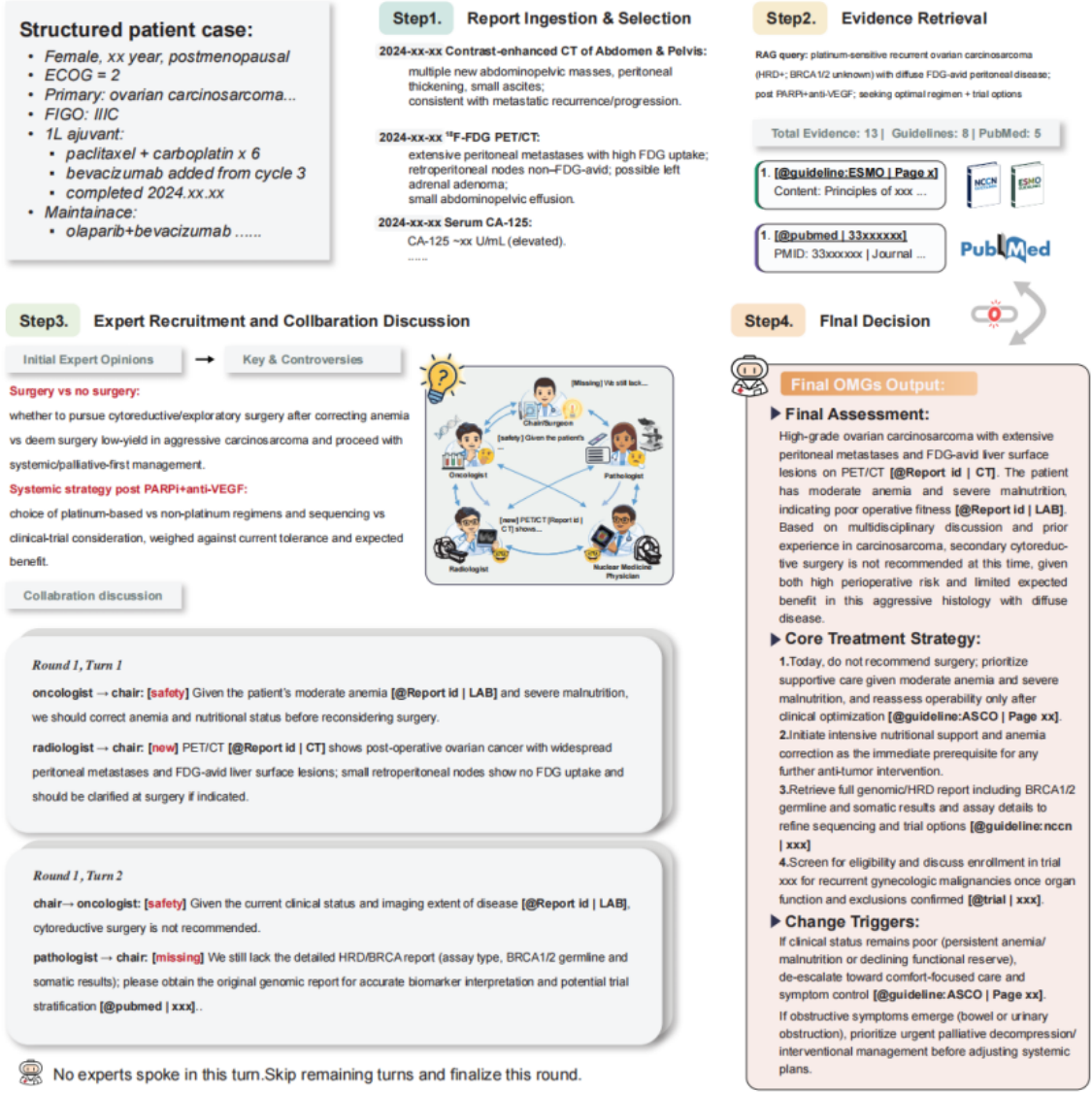

**Extended Data Fig. 1 | Illustrative example of OMGs in an ovarian tumour MDT case**

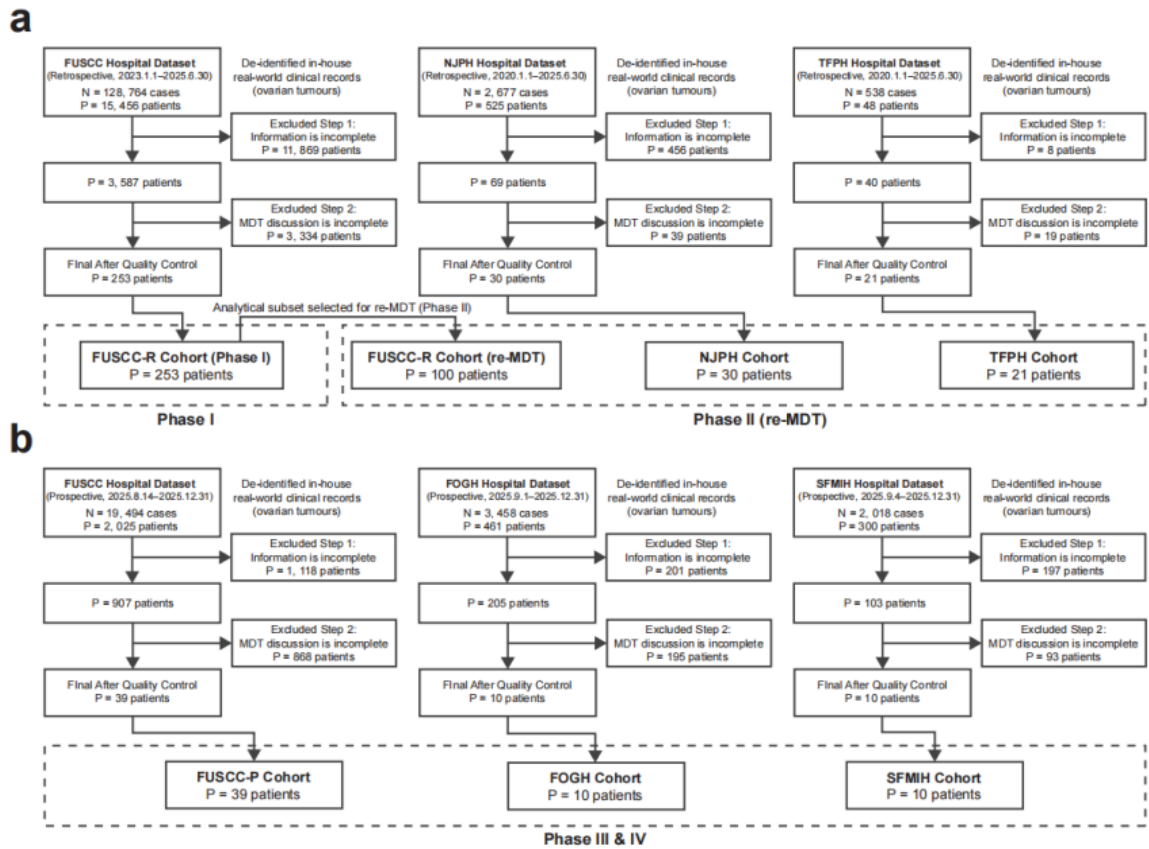

**Extended Data Fig. 2 | Cohort screening, exclusions and final analytic samples for the retrospective (Phases I–II) and prospective (Phases III–IV) multicentre cohorts.**

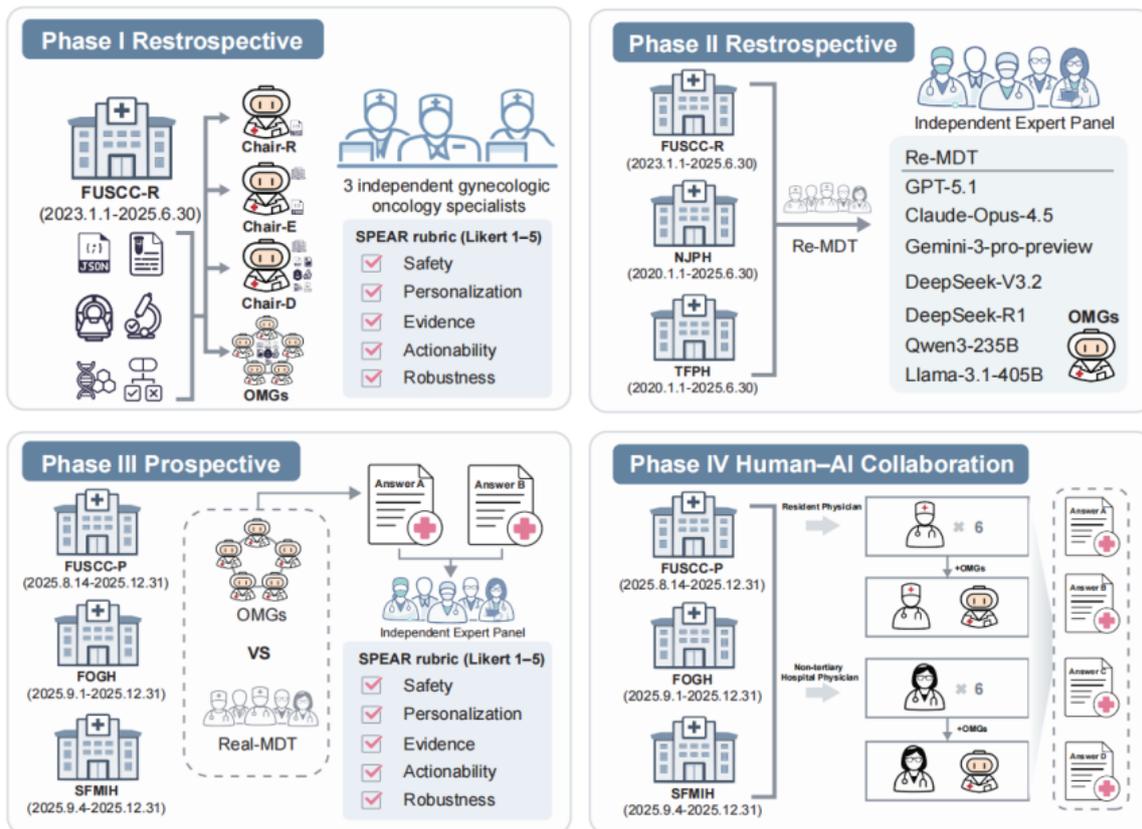

**Extended Data Fig. 3 | Four-phase evaluation design for OMGs.**

Phase I, single-centre retrospective benchmarking against progressively augmented single-agent baselines; Phase II, multicentre retrospective evaluation across large language model backbones and against a re-MDT panel; Phase III, prospective comparison between OMGs and routine MDT conclusions; Phase IV, paired human–AI collaboration assessed using the SPEAR rubric (Likert 1–5).

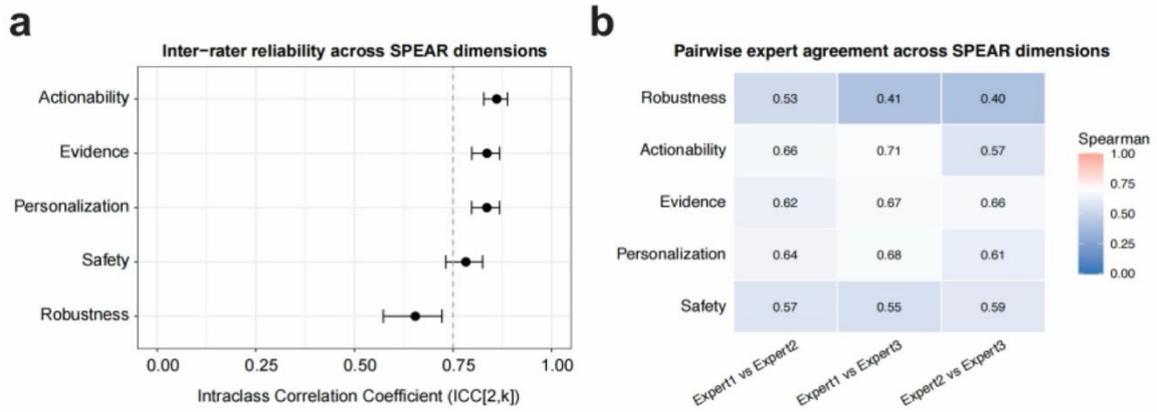

**Extended Data Fig. 4 | Inter-rater agreement of expert SPEAR ratings in Phase I.**

**a**, Dimension-level inter-rater reliability estimated using ICC(2,k) with 95% confidence intervals (dashed line indicates ICC=0.75); **b**, pairwise Spearman correlations between experts across the five SPEAR dimensions.